\documentclass[letterpaper, 10 pt, conference]{ieeeconf}  

\IEEEoverridecommandlockouts                              

\usepackage{amsmath}
\usepackage{graphicx}
\usepackage[colorlinks=true, allcolors=blue]{hyperref}
\usepackage{subcaption}
\usepackage{tabularx}

\title{\LARGE \bf
ParaPose: Parameter and Domain Randomization Optimization for Pose Estimation using Synthetic Data
}

\author{Frederik Hagelskjær and Anders Glent Buch
\thanks{This project was funded in part by Innovation Fund Denmark through the project MADE FAST, in part by the SDU I4.0-Lab.}
\thanks{
Both authors are with SDU Robotics, Mærsk Mc-Kinney Møller Institute, University of Southern Denmark, 5230 Odense M, Denmark
        {\tt \{frhag,anbu\}@mmmi.sdu.dk}}%
}

\begin{document}

\maketitle
\thispagestyle{empty}
\pagestyle{empty}

\begin{abstract}

Pose estimation is the task of determining the 6D position of an object in a scene. Pose estimation aid the abilities and flexibility of robotic set-ups. However, the system must be configured towards the use case to perform adequately. This configuration is time-consuming and limits the usability of pose estimation and, thereby, robotic systems.

Deep learning is a method to overcome this configuration procedure by learning parameters directly from the dataset. However, obtaining this training data can also be very time-consuming. The use of synthetic training data avoids this data collection problem, but a configuration of the training procedure is necessary to overcome the domain gap problem. 
Additionally, the pose estimation parameters also need to be configured. 
This configuration is jokingly known as grad student descent as parameters are manually adjusted until satisfactory results are obtained.

This paper presents a method for automatic configuration using only synthetic data. This is accomplished by learning the domain randomization during network training, and then using the domain randomization to optimize the pose estimation parameters. The developed approach shows state-of-the-art performance of 82.0~\% recall on the challenging OCCLUSION dataset, outperforming all previous methods with a large margin. These results prove the validity of automatic set-up of pose estimation using purely synthetic data.

\end{abstract}

\section{INTRODUCTION}

The pose of a rigid object can be represented as a 6D transform consisting of translation and rotation. This 6D pose allows a robotic set-up to manipulate and analyze the object. By obtaining this pose through visual pose estimation, the flexibility of the set-up is greatly increased.
Without visual pose estimation, the position of objects needs to be known beforehand by using mechanical fixtures. Introducing new objects thus requires the assembly of new fixtures.

\begin{figure}[tb]
    \centering
    \begin{subfigure}[t]{.24\textwidth}
      \centering
      \includegraphics[trim=350 50 350 70, clip, width=0.99\linewidth]{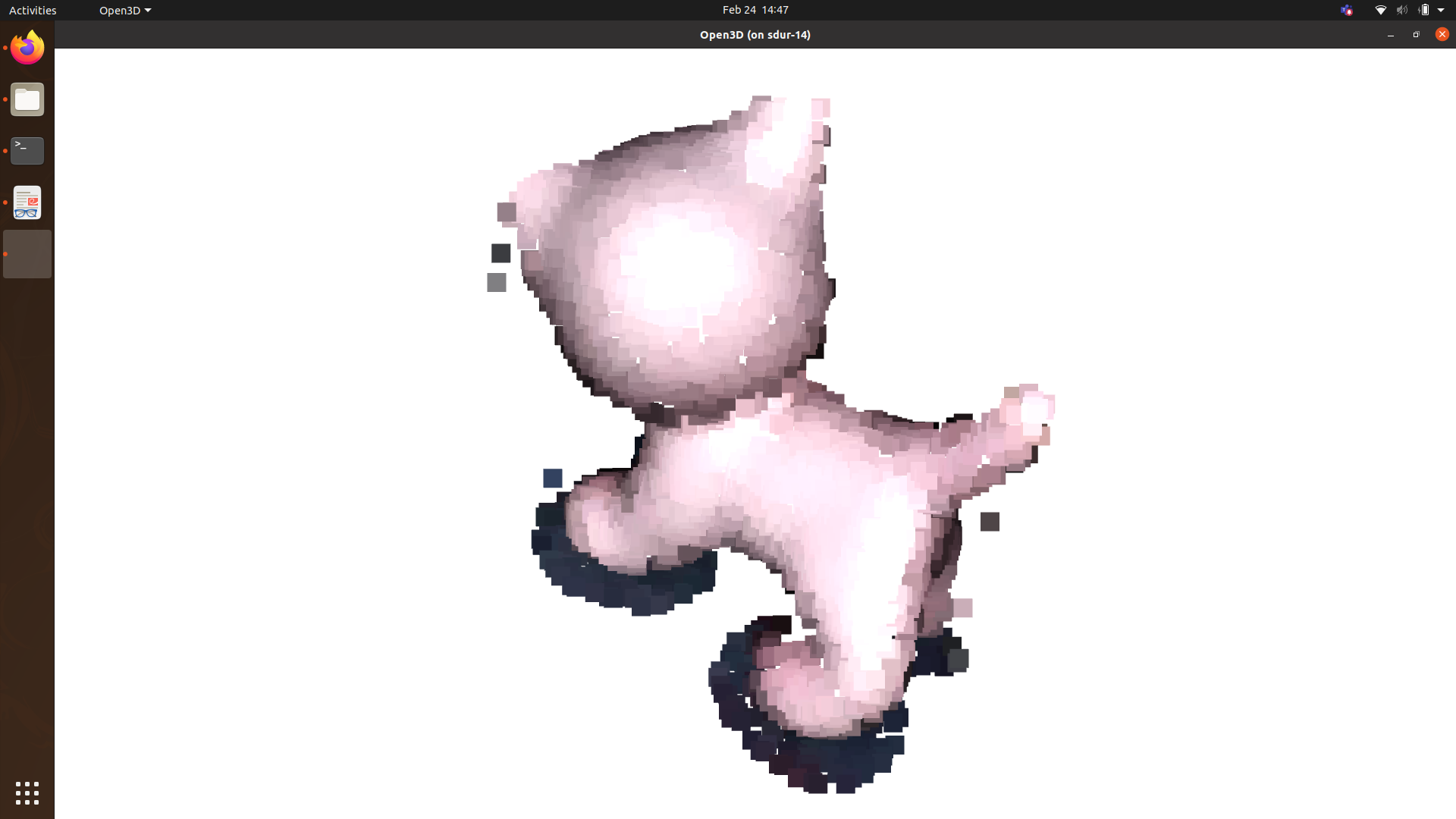}
      \caption{Synthetic training data}
      \label{fig:res:synth}
    \end{subfigure}%
    ~
    \begin{subfigure}[t]{.24\textwidth}
      \centering
      \includegraphics[trim=350 50 350 70, clip, width=0.99\linewidth]{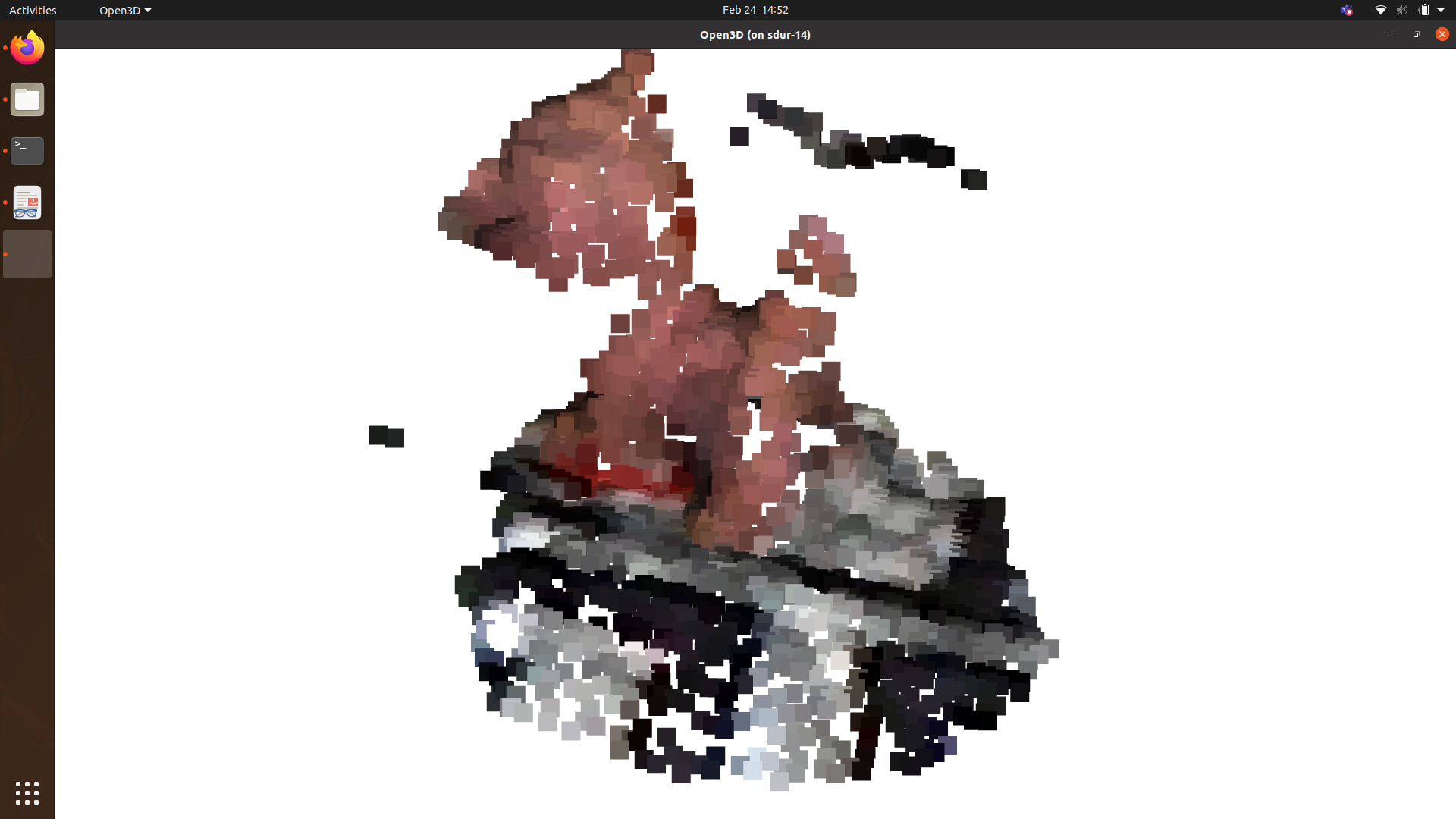}
      \caption{Real test data}
      \label{fig:res:real}
    \end{subfigure}%

    \begin{subfigure}[t]{.24\textwidth}
      \centering
      \includegraphics[trim=350 50 350 70, clip, clip, width=0.99\linewidth]{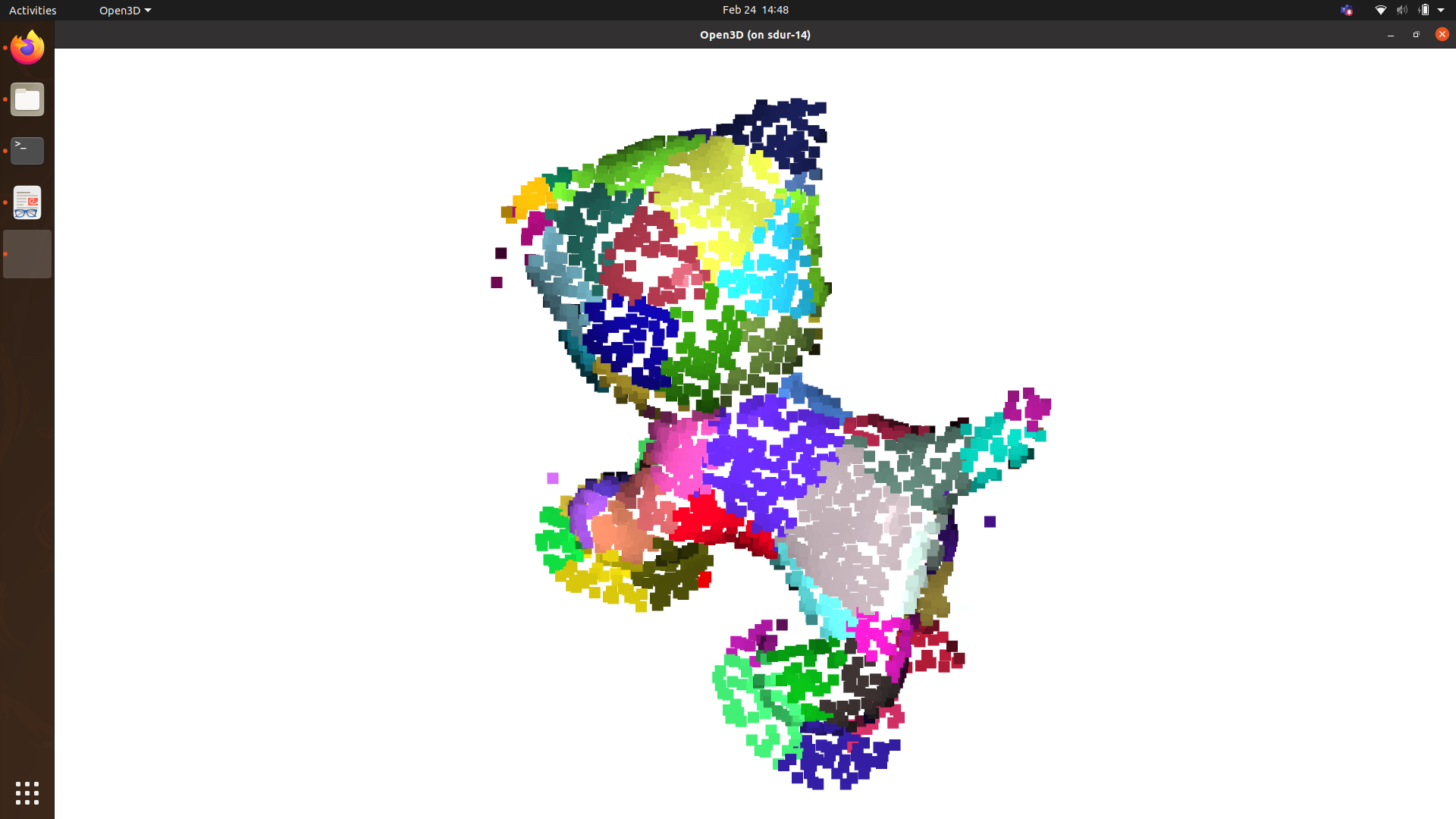}
      \caption{No DR}
      \label{fig:res:synth_pred_no_noise}
    \end{subfigure}%
    ~
    \begin{subfigure}[t]{.24\textwidth}
      \centering
      \includegraphics[trim=350 50 350 70, clip, clip, width=0.99\linewidth]{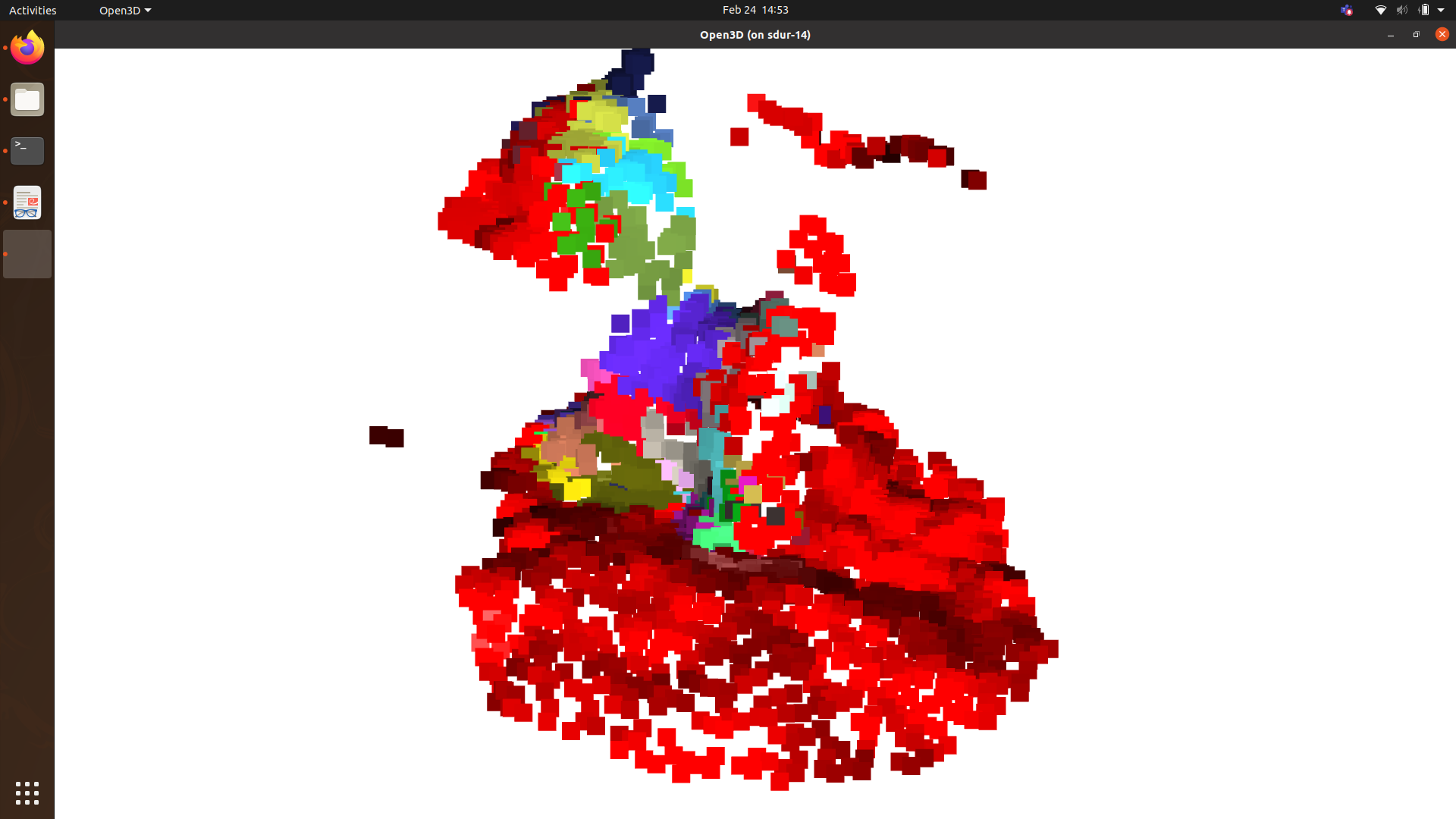}
      \caption{No DR}
      \label{fig:res:real_pred_no_noise}
    \end{subfigure}%
    
    \begin{subfigure}[t]{.24\textwidth}
      \centering
      \includegraphics[trim=350 50 350 70, clip, clip, width=0.99\linewidth]{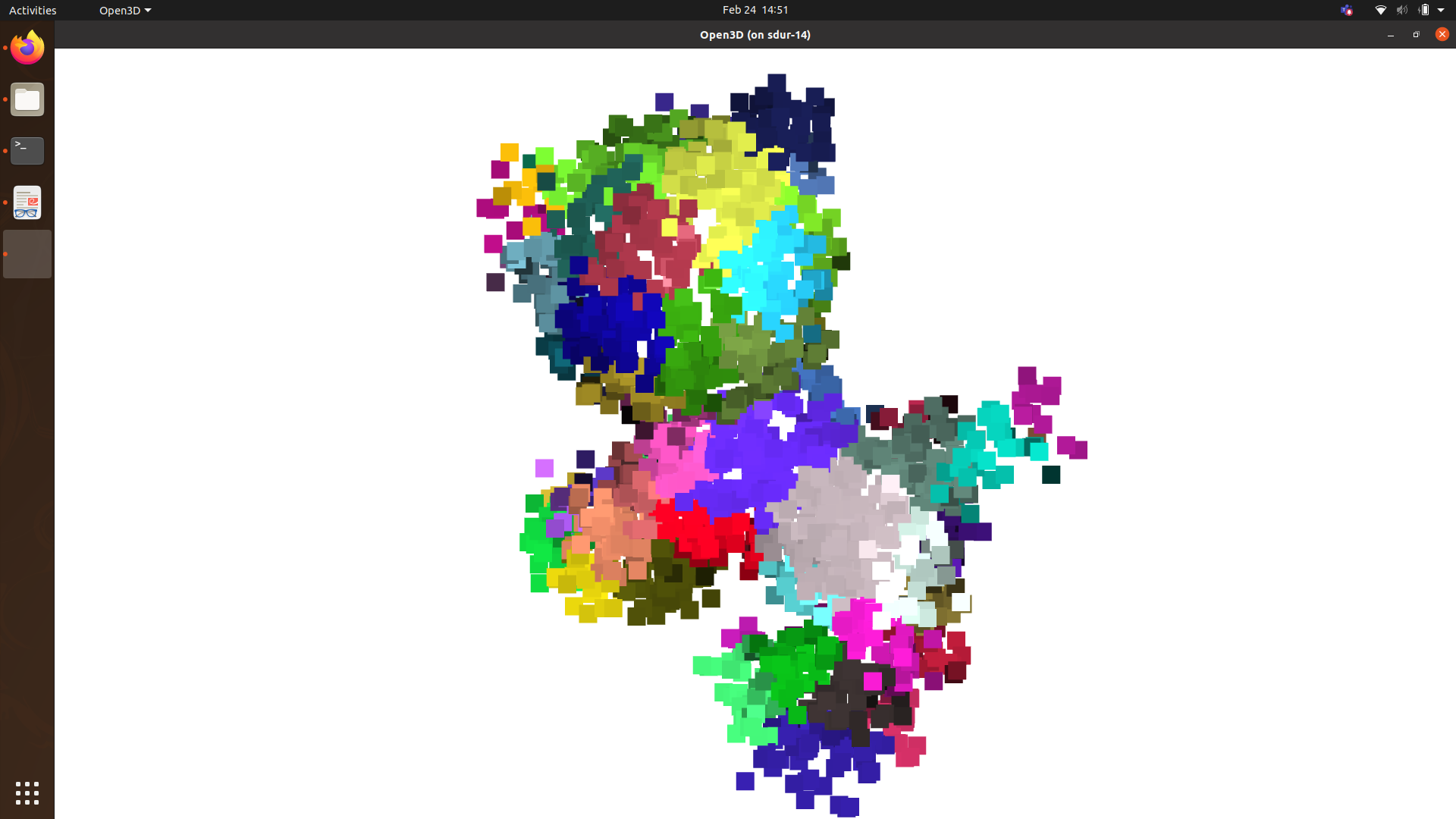}
      \caption{With DR}
      \label{fig:res:noise_pred}
    \end{subfigure}%
    ~
    \begin{subfigure}[t]{.24\textwidth}
      \centering
      \includegraphics[trim=350 50 350 70, clip, clip, width=0.99\linewidth]{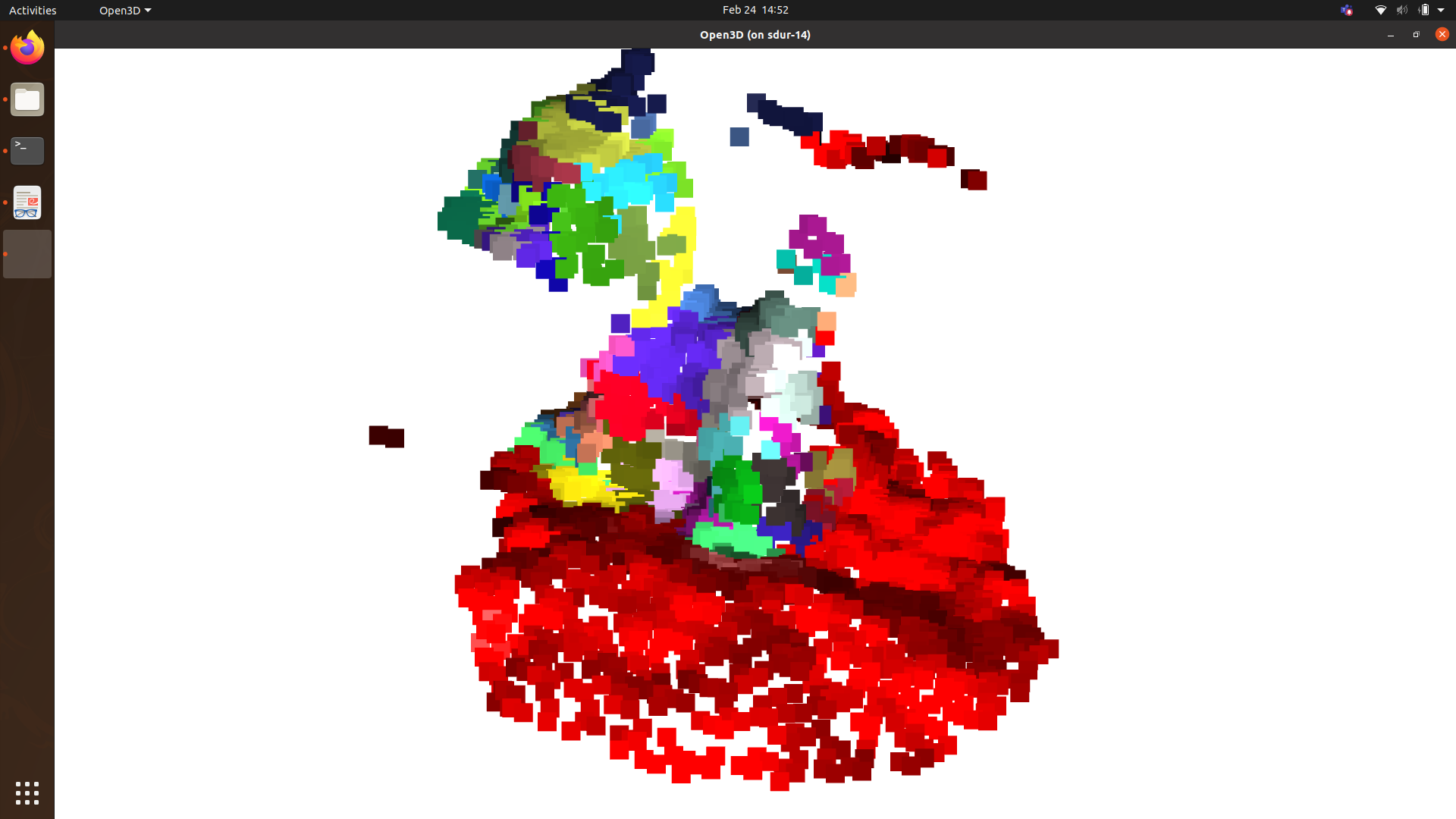}
      \caption{With DR}
      \label{fig:res:real_pred_noise}
    \end{subfigure}%
    
    \caption{ Visualization of the domain randomization's (DR) impact on the network's generalization ability. (a) and (b), show synthetic and real data, respectively.
    (c) show a perfect feature prediction on synthetic data without domain randomization. In (d), the network trained without domain randomization shows bad results on the real test data. (e) show the more difficult feature prediction with domain randomization, but increased performance on the test scene in (f).
    }
    \label{fig:res}
\end{figure}

%
However, the complex configuration of the visual pose estimation system can make the set-up very time-consuming. 
Different scenarios vary greatly, and it is essential to fit the configuration to the scenario. Variations such as camera type, lighting, background clutter, and levels of occlusion all influence the system. 
%
While controlling these influences can simplify the set-up \cite{hagelskjaer2019using}, the environment cannot always be controlled. For example, attempting to remove shadows in an image can quickly become a time-consuming task.
Properties such as object material and shape will also influence the performance of the pose estimation algorithm.
In academia, good results are often obtained by the method known as Grad Student Descent \cite{gencoglu2019hark}, where hyper-parameters are continuously tuned until adequate results are obtained.

As a result of the difficulties, the set-up time can often last more than a week \cite{hagelskjaer2017does}. This set-up time significantly reduces the usability of robotics. And as most of the time is spent on adjusting the software \cite{hagelskjaer2017does}, it is essential to simplify this part.
%
%
One approach has been the automatic tuning of pose estimation parameters using training data \cite{hagelskjaer2019bayesian}. This method uses real data to train the parameters. This approach has even shown increased performance compared with parameters tuned manually. 
Deep learning algorithms have used this approach very successfully, using real training data to tune the network structure. On most benchmarking datasets, the top-performing algorithms are based on deep neural networks \cite{hinterstoisser2012model, brachmann2014learning, kaskman2019homebreweddb, xiang2017posecnn}.
However, obtaining real training is a task that needs to be performed manually. Moreover, for deep learning models, this can easily amount to more than hundreds of images that must be collected and correctly labeled.
To overcome this obstacle, the use of synthetic training data has been introduced \cite{thalhammer2019towards, denninger2019blenderproc, kehl2017ssd}. By using synthetic training data, the manual set-up is greatly reduced, while the algorithm retains high performance. 
In the BOP challenge \cite{hodan2020bop} methods using synthetic training data obtain results in the same range as methods trained on real data. 
The method that the optimization in this paper is based on also obtains results on the LINEMOD \cite{hinterstoisser2012model} and OCCLUSION \cite{brachmann2014learning} datasets similar to methods trained on real data.
However, the success of methods trained on synthetic data is highly dependent on domain randomization \cite{hagelskjaer2020bridging}, which is configured manually. The pose estimation parameters are also found heuristically to obtain the best results.


In this paper, we present a method for the automatic optimization of pose estimation parameters. The optimization is performed using the same synthetic data used for training the deep learning model. However, our results show that domain randomization is necessary during parameter optimization to obtain good results. The effect of the domain randomization is shown in Fig.~\ref{fig:res}. To this end, we employ the domain randomization from the deep learning training. Using domain randomization drastically improves the performance and allows our model to outperform models trained with real data and heuristically found parameters. On the challenging OCCLUSION \cite{brachmann2014learning} dataset, our method trained with synthetic data, and optimized with synthetic data, obtains a recall of 82.0~\%, which is the new state-of-the-art performance.

To create a usable parameter optimization system, the parameters are split into two groups. This allows the isolation of parameters that both influence performance and run-time. For many applications, this run-time is very important with respect to the feasibility of the system. Our optimization allows the user to obtain the best performance to the desired maximum run-time.

This paper presents the following main contributions.
\begin{itemize}
    \item A method for the automatic set-up of pose estimation systems using only synthetic data.
    \item A simple method to obtain the desired run-time of the system.
    \item State-of-the-art results on the challenging OCCLUSION dataset.
\end{itemize}
The remaining paper is structured as follows: We first review related papers in Sec.~\ref{related}. In Sec.~\ref{pose_estimation}, our developed method is explained. In Sec.~\ref{evalution}, the parameter optimization is performed, and the performance is verified. Finally, in Sec.~\ref{conclusion}, a conclusion is given to this paper, and further work is discussed.

\section{RELATED WORK}
\label{related}

Since the introduction of deep neural networks, domain randomization has always been present \cite{shorten2019survey}. Domain randomization is used to minimize the distance between the training and test sets. The less the training set represents the test set, the more critical domain randomization is. 

The most common domain randomization strategies are adding Gaussian noise, randomly cropping the image, rotating the image, and changing, hue, contrast and brightness \cite{shorten2019survey}. These methods are generally seen in many pose estimation methods \cite{kehl2017ssd, peng2019pvnet, he2020pvn3d}.

There also exist other strategies for the generalization of networks. One commonly used strategy is pre-training, where the network is trained on a large and diverse dataset to learn features, which are then re-used when trained on the actual dataset. Another approach is the dropout of features during training to avoid the network relying on a single feature. 


\subsubsection{Pose Estimation}

Successful pose estimation methods are generally based on networks operating on 2D images. This is a result of the success of 2D convolutional networks, as many good networks have been developed with good pre-trained weights and domain randomization strategies \cite{he2017mask}. The following pose estimation methods all use 2D networks for pose estimation. 3D information is then often used to refine the poses. In our approach, we use the 3D data directly in the learning. It gives the advantage that all modalities are included in the network, but the disadvantage is that no networks pre-trained on large networks are available.

An example is SSD-6D \cite{kehl2017ssd}, a deep learning based pose estimation method using only synthetic data. Renders of objects are cut and pasted onto images from the COCO \cite{lin2014microsoft} dataset. The network is trained to classify the rotation and presence of objects. Domain randomization is performed by changing the brightness and contrast. The method is tested on the LINEMOD dataset, where it obtains good performance but is outperformed by methods trained with real data. 
In PVNet \cite{peng2019pvnet} 10000 images are rendered and cut and pasted onto images from the SUN397 \cite{xiao2010sun} dataset. Then random cropping, resizing, rotation, and Gaussian noise is applied. This network also uses real training data along with synthetic data.
PVN3D \cite{he2020pvn3d} use the training data and domain randomization from PVNet \cite{peng2019pvnet}, but expand with 3D point-clouds. The network obtains very good performance on the LINEMOD \cite{hinterstoisser2012model} dataset, which it is trained on, but on the more challenging OCCLUSION dataset, it does not perform well.
PoseCNN \cite{xiang2017posecnn} is the current state-of-the-art method on the OCCLUSION \cite{brachmann2014learning} dataset. Real data is used along with synthetic renders with objects placed randomly in a scene.  
The DPOD \cite{zakharov2019dpod} method uses real training data and object renderings which are cut and pasted onto images from the COCO \cite{lin2014microsoft} dataset. Brightness, saturation, and Gaussian noise is used to increase the generalizability. 
Similar to our method DenseFusion \cite{wang2019densefusion} process point-cloud data. The point-cloud processing is combined with 2D processing, and the network is only trained on real data. The network is tested on the LINEMOD \cite{hinterstoisser2012model} and obtains good results, but is outperformed by methods using both using synthetic \cite{hagelskjaer2020bridging} or real data \cite{he2020pvn3d}.
%
The method CosyPose \cite{labbe2020cosypose} show very good results with extensive domain randomization. One million synthetic training images are generated, with objects placed randomly. Gaussian blur, contrast, brightness, color, and sharpness filters are all added to the images. The cut and paste strategy are then used to increase the background variance. This data, along with real training data, show state-of-the-art performance on two pose estimation datasets. The method also showed the best recall for methods using only synthetic data on the 2020 BOP challenge \cite{hodan2020bop}.
As mentioned, most of these methods only use the cut and paste methods to generate the background. The BlenderProc \cite{denninger2019blenderproc} dataset use photo-realistic renderings to include reflections and shadows. However, methods using this dataset still use the cut and paste methods to increase performance \cite{labbe2020cosypose}.

\subsubsection{Learning Domain Randomization}

In the reviewed pose estimation methods, the domain randomizations are manually tuned. However, a number of methods exist to tune the domain randomization automatically. 
One approach is to directly minimize the domain gap between real and synthetic data by introducing unlabelled real data \cite{ganin2016domain}. During training on the synthetic data, the learned features should not be able to discriminate between real and synthetic data.

The domain randomization problem can also be seen as an optimization task \cite{zoph2020learning}. Here the COCO \cite{lin2014microsoft} dataset is used as a validation set where different domain randomization strategies are tested. The best domain randomization policy is then used for testing.
The same strategy is seen in AutoAugment \cite{cubuk2019autoaugment} which tests different domain randomization strategies on a validation dataset. Here state-of-the-art results are obtained on the many datasets.

Domain randomization can also be used to increase robustness against adversarial images \cite{yu2019pda}. Similar to our method, the goal is to increase the amount of domain randomization instead of finding the correct domain randomization. They show that this improves performance compared with fixed domain randomization.

\subsubsection{Domain randomization in 3D}

Successful deep learning of point-clouds was introduced with PointNet 
\cite{qi2017pointnet}. Here domain randomization consists of random Gaussian noise in the overall position, Gaussian noise for each point, and a random rotation. We base our domain randomization on this approach but add random Gaussian noise to the two added channels of normal vector and RGB color. Additionally, as the image is obtained with a 2.5D scanner, we add flattening of the point-cloud to simulate wrongly obtained depth.

For 3D point-clouds a number of new domain randomization methods have also been developed. 
One approach is cross-modal domain randomization \cite{wang2021pointaugmenting}, where the object is cut and pasted into different scenes. More similar to the approach seen in 2D, where more samples can be created from small amounts of data.
Another method to create more samples from existing data is point interpolation \cite{chen2020pointmixup}. New data is generated by interpolating between the points from two different point-clouds. These two methods are useful when only small amounts of data are present. 
Learning domain randomization for point-clouds has also been performed. In PointAugment \cite{li2020pointaugment} a network is trained to generate domain randomization. The domain randomization is both point-wise and shape-wise. The augmentation is optimized during network training with the requirement that the loss with augmentation should be slightly larger than without augmentation. This approach shows improved performance
compared with classic augmentation. 
A method that successfully uses 3D domain randomization is PointVoteNet \cite{hagelskjaer2019pointvotenet}, which is trained on real data. To create more samples background data is cut and pasted.
%

\subsubsection{Previous Method}

Our pose estimation method is based on a previous method \cite{hagelskjaer2020bridging}. The method is trained on the BlenderProc \cite{denninger2019blenderproc} dataset created for the BOP challenge \cite{hodan2020bop}. This method use domain randomization based on the Kinect sensor. In this paper, the domain randomization is optimized during the training without knowledge of the scene.

An example of optimizing classic computer vision is seen with Bayesian Optimization \cite{hagelskjaer2019bayesian}. The parameters of an existing pose estimation method are optimized using real data, and the improvement gives state-of-the-art performance. We use a similar approach in our paper, but synthetic images with domain randomization are used instead of real training data. Thus, the set-up of our method is entirely based on synthetic data.









\section{METHOD}
\label{pose_estimation}


The pose estimation in this paper is based on an existing method \cite{hagelskjaer2020bridging}. In the first step, a candidate detector is used, and from this output, the method extracts point-clouds. These point-clouds are fed through a modified DGCNN \cite{dgcnn} network, and three predictions are obtained; classification, background segmentation, and votes. The classification is the probability that the object is in the center of the point-cloud. This is used to filter point-clouds, as performing pose estimation for each point-cloud is very time-consuming. The background segmentation and the votes are combined for the remaining point-clouds to create keypoint matches to the object model. RANSAC  \cite{fischler1981random} is then used to find possible pose estimations, which are further refined by Coarse to Fine Iterative Closest Point (C2F-ICP). Finally, the best pose estimation is determined by a depth check using the projected model.
Compared with the previous method \cite{hagelskjaer2020bridging}, the RANSAC, ICP, and depth distances are scaled by object diagonal. The depth check is also aided by a contour check.

The deep neural networks are trained using only synthetic data from the BlenderProc dataset \cite{hodan2020bop, denninger2019blenderproc}. However, all parameters in the pose estimation algorithm were originally found heuristically. In this paper, these parameters are instead found by optimization, using the same synthetic data.


\subsection{Candidate Detector}

In the original implementation, a MASK R-CNN \cite{he2017mask} trained on the synthetic BlenderProc \cite{denninger2019blenderproc} dataset was used as the candidate detector. However, the network's ability to classify point-clouds makes it agnostic to the candidate detector. As a result, the method is also tested using detections from the CosyPose method \cite{labbe2020cosypose}. CosyPose was selected based on the good results using synthetic data on the BOP datasets \cite{labbe2020cosypose}. The detections have been created by running the CosyPose detector trained on the BlenderProc dataset on the complete LINEMOD \cite{hinterstoisser2012model} and OCCLUSION \cite{brachmann2014learning} datasets.

\subsection{Domain Randomization Optimization}

\begin{figure}[tb]
    \vspace{1.5mm}
    \centering
    \includegraphics[trim=0 0 0 0, clip, width=0.95\linewidth]{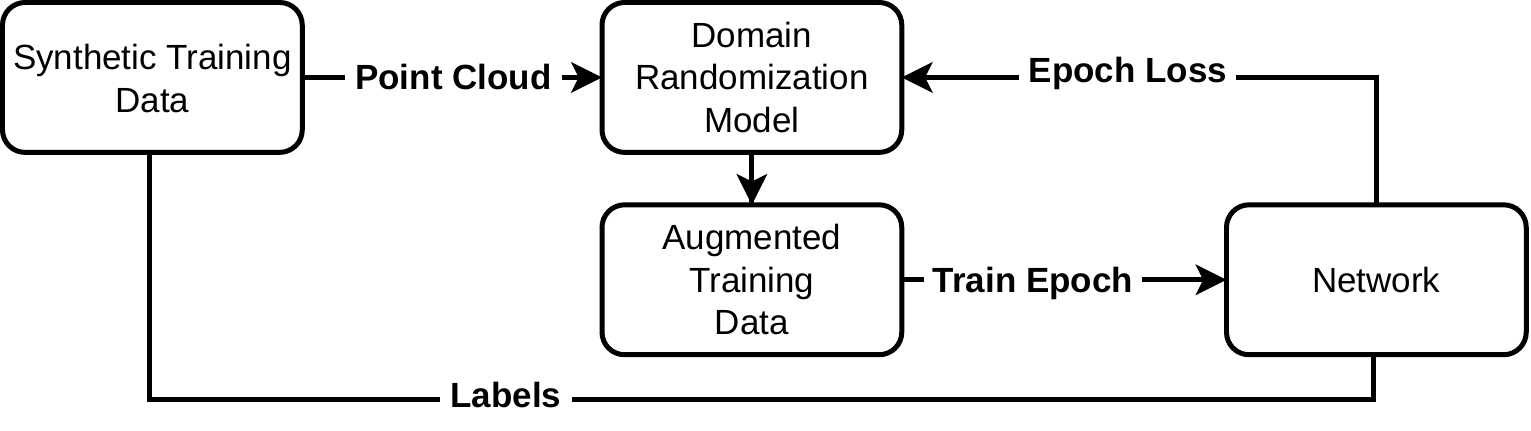}
    \caption{Overview of the network training and domain randomization optimization. The domain randomization model is updated after each epoch.}
    \label{fig:model_training}
    \vspace{-6mm}
\end{figure}

The method is based on a subset of domain randomizations, each with a noise level. As in other methods \cite{peng2019pvnet}, the level of noise in each sample is a random Gaussian based on a selected max level. 
The task of the optimization is to determine the max noise level for each subset of domain randomization, which best improves the generalizability.

The six subsets of domain randomization are Gaussian XYZ noise, Gaussian normal vector noise, Gaussian RGB noise, Gaussian RGB shift, rotation, and flattening.
The noise starts at a predetermined level of 1.0, 0.02, 0.02, 0.04, 5.0, and 0.02, respectively. Where point noise is in millimeter, rotation is in degrees, and all other is in percentage. For each parameter, the jump size is half the original value.

The first four epochs are run without domain randomization. After which, the domain randomization is activated, with noise at predetermined levels. At eight epochs, the optimization of the domain randomization is started by recording the loss level and increasing the first noise level.
After one epoch, the loss is measured and compared with the recorded loss. If the loss is decreased by more than 2.5 percent, then the noise isn't limiting the training, and the noise level is increased again. Otherwise, the next noise type is increased. If the loss increase is more than five percent, the max level is reduced to the previous level, and this noise type is no longer increased. An overview of the training and domain randomization procedure is shown in Fig.~\ref{fig:model_training}.

\subsection{Parameters}

Several parameters are tuned in this pose estimation method to obtain good performance. To explain the method, an overview is given of the optimized parameters. For a further explanation of the method, the reader is referred to the original paper \cite{hagelskjaer2020bridging}.
We divide the parameters into continuous and discrete parameters. The continuous parameters can have any real value within a known range. An important aspect is that fine-tuning these values will primarily increase the precision without influencing the run-time. 
The discrete parameters are integer values that impact both the precision and run-time of the algorithm. It is, therefore, essential to find the optimal trade-off for the given task.

For the discrete parameters, the following parameters are optimized.
The number of point-clouds which are classified, $PC$. The remaining point-clouds to perform pose estimation upon, $PE$. The number of RANSAC iterations, $RI$. The number of RANSAC proposals to be sorted with the depth check, $DC$. And the number of ICP iterations, $II$.

For the continuous parameters, the following are optimized. 
The cut-off radius size to determine if less than 2048 points are accepted, $sr$.
The threshold for accepting votes at symmetric positions, $vt$. The distance in mm for the RANSAC algorithm, $rd$. The C2F-ICP distance in mm, $id$. The scaling in the C2F-ICP, $is$. The distance in mm to determine whether a point belongs to foreground or object, $bd$. The distance for accepting object points, $ad$.

Additionally, a number of parameters are seen as essential to the structure of the algorithm and are kept fixed. The C2F-ICP has three resolutions. The network structure has 2048 input points. A minimum of 512 points is needed to process a point-cloud, and a minimum of 100 key-point matches are necessary for the RANSAC algorithm. The point-cloud is pre-processed to a 1mm voxel grid. The object is pre-processed to a 5mm voxel grid for the C2F-ICP. The object model is maximally at 100 keypoints on object model. The RANSAC iterations are split into parts of 50. Normal vectors are computed with a radius of 10 mm.

\subsection{Pose Estimation Parameter Optimization}

As stated in Sec.~\ref{pose_estimation} two different types of parameters are used in the pose estimation method. The discrete parameters have a direct influence on the run-time, while the continuous parameters do not. As a result of these differences, two different parameter optimizations are performed.

First, the continuous parameters are found with fixed discrete parameter values, and the optimized continuous parameters are then used for the discrete optimization.

\subsubsection{Continuous Parameters}

For the continuous parameters, the goal is to find the optimal parameter set for performance. As many parameters are present, the parameters are expected to influence each other. The risk of a local optimization finding a local maximum is, therefore, present. To avoid this, a global optimization method is used. 
The chosen optimization method is Bayesian Optimization with Upper Confidence Bound as the acquisition function \cite{snoek2012practical}. This method has been used successfully for pose estimation parameter optimization \cite{hagelskjaer2019bayesian}. It can be represented by a single parameter, which is the trade-off between exploration and exploitation. The method works without residuals, which is useful as very small changes in parameter values will generally not change the recall. It is a bounded optimization, and most values have logical bounds, and the remaining can be bounded by reason, such as metric bounds.

The optimization consists of four iterations, with the first being random exploration and then moving to an exploitation strategy. This is achieved by decreasing the $\kappa$ value. 50 random iterations are performed, followed by 100 with $\kappa = 0.5$, then 50 with $\kappa = 0.1$, and finally 50 with $\kappa = 0.01$, resulting in 250 iterations in all.

\subsubsection{Discrete Parameters}

The discrete parameter values both influence the recall and the run-time. As a result, one cannot simply determine the best parameter set.
Instead, the goal of this optimization is to find a list of parameter sets. This list will give the parameter sets that give the highest increase in recall when increasing the run-time. 
A user would then be able to select the best trade-off for the given situation. 

As the parameter values are discrete, the optimization is performed with a grid search. All parameter sets are sorted according to run-time, and starting with the lowest run-time, parameter sets are added if they increase the recall.

\subsubsection{Varying Number of Objects}

The algorithm's run-time depends not only on the parameter values but also on the number of searched objects. While a new optimization could be performed if any objects are removed or added, it is desirable to perform the optimization once and then adjust the parameters to fit run-time given the number of objects. The relationship between discrete parameter values, number of objects, and run-time are shown in Eq.~\ref{eqn:runtime}. This equation can be solved using least-squares on the grid search results. Here $t_{image}$ is the run-time of the system, $t_{pre}$ is the pre-processing and initial detection of interest points,
$t_{net}$ is the is run-time of the network, $t_{ran}$ is the run-time of a RANSAC iteration, $t_{icp}$ is the run-time of an ICP iteration, and $t_{depth}$ is the run-time of the depth check.

\begin{equation}
\label{eqn:runtime}
   \begin{aligned}
   t_{image} = t_{pre} +  obj * ( t_{net} * PC + PE * \\ ( t_{ran} * RI + DC * ( t_{icp} * II + t_{depth} )))
   \end{aligned}
\end{equation}

\subsection{Evaluation Metrics}
The success criteria of a pose estimation must first be defined before the performance can be optimized. 

A simple score is the distance between the ground truth pose and the pose estimate. This is performed in the Tejani \cite{tejani2014latent} and Doumanoglou \cite{doumanoglou2016recovering} datasets poses with less than 50mm error are declared correct. This metric also adds a fifteen-degree maximum error between the Z-axes to include angular errors while accommodating symmetric objects.
Two central problems with this score are that errors further away from the object center are penalized less severely, and only angular errors in the Z-axis are included.
%
For the LINEMOD \cite{hinterstoisser2012model} and OCCLUSION \cite{brachmann2014learning} datasets, the average distance (ADD) metric is used. First, the object model is converted to a point-cloud which is projected into the scene both by the ground truth and the found pose. The ADD metric is then computed as the average distance between each point in the two point-clouds. A pose estimate is successful if the average distance is smaller than ten percent of the object model's diagonal. The ADD/I score can be used to accommodate symmetric objects where the distance is simply to the nearest point in the comparison cloud.
As the ground truth labels in the OCCLUSION dataset are not very precise, this metric fits well with the dataset. And as the test datasets use this score for evaluation, it would seem obvious to use this for the optimization. 
However, as the parameter optimization is performed with synthetic data,  the ground truth is exact, and a more discriminating metric can be used. Therefore, the metric established from the BOP test is used for the optimization. 
The metric consists of three different scores, Visible Surface Discrepancy (VSD) \cite{hodan2018bop,hodavn2016evaluation}, Maximum Symmetry-Aware Surface Distance (MSSD) \cite{drost2017introducing}, and Maximum Symmetry-Aware Projection Distance (MSPD) \cite{hodan2020bop}. To understand the metric a very general description is given for each score. 
The object is first projected into the scene according to the ground truth pose and the pose estimate. In VSD, the distance in depth, for the visible part of the model is calculated. In MSSD, the maximum distance between each point in the model is calculated. The last score, MSPD, is the distance in pixels of the projected object and is therefore good for methods without depth data.
Multiple thresholds are used instead of a single cut-off value, and the multiple scores are combined into an average.

\section{EXPERIMENTS}
\label{evalution}

To show the performance of the developed method, tests are performed on the LINEMOD dataset \cite{hinterstoisser2012model} and the challenging OCCLUSION \cite{brachmann2014learning} dataset. The parameters are first optimized, and the optimized method is then tested on the real data and compared with the previous approach and current state-of-the-art methods.
The optimization is performed with the synthetic BlenderProc dataset \cite{hodan2020bop,denninger2019blenderproc}. For the network training, 10000 synthetic images are used, and for the parameter optimization, nine previously unseen synthetic images are used. All experiments are performed on a PC environment (Intel i9-9820X 3.30GHz CPU and NVIDIA GeForce RTX 2080 GPU). 

\subsection{Network and Domain Randomization Optimization}

\begin{figure}[t]
    \vspace{1.5mm}
    \centering
    \includegraphics[trim=0 0 0 0, clip, width=0.95\linewidth]{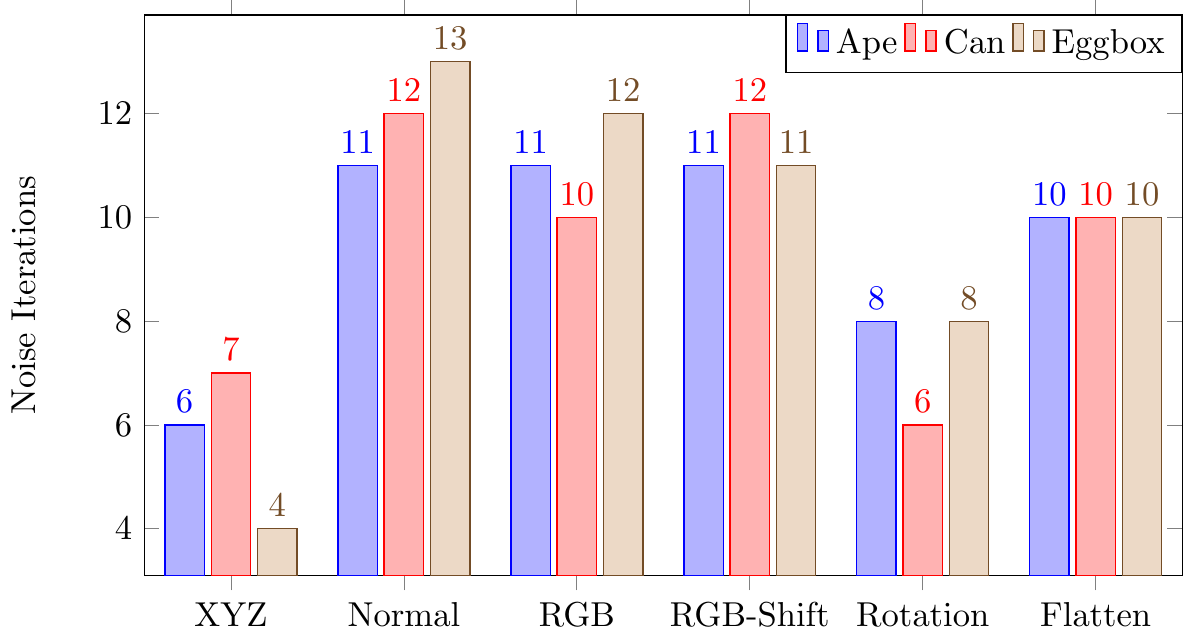}
    \caption{Number of noise level increases for each noise type shown for three different types of objects. }
    \label{fig:noise_opt_graph}
\end{figure}

For each object, a network and domain randomization is trained. The training runs for sixty epochs with parameters according to the previous method \cite{hagelskjaer2020bridging}. The resulting noise levels for three objects are shown in Fig.~\ref{fig:noise_opt_graph}. The objects are Ape, Can, and Eggbox, as they respectively represent small, large, and symmetric objects. It is seen that the network tolerates a much lower noise increase for XYZ and rotation. The noise levels appear similar for all objects, although the symmetric object Eggbox has a very low XYZ level.

\subsection{Continuous Optimization}

The continuous optimization is performed using Bayesian Optimization. The discrete parameters are fixed to values of, $PC=32$, $PE=6$, $RI=500$, $II=10$, and $DC=2$. By setting $PE=6$ and $DC=2$, twelve pose estimations are checked for each object. This ensures that all the continuous parameters are used for the pose estimation without making the run-time unfeasible. The optimization is performed both with and without domain randomization. 
Additionally, an optimization is made using the ADD/I metric instead of the BOP.
The algorithm's recall resulting from the parameters span from 40~\% to 93~\%. The correct parameters thus have a strong influence on the performance. The parameters with and without domain randomization, and the heuristically found parameters, are shown in Tab.~\ref{tab:found_con}.
%
Compared with the heuristic parameters, a notable difference is seen with the much lower voting threshold, $vt$, giving more matches to RANSAC.
Compared with the optimization without domain randomization, the C2F-ICP and depth check parameters are the most significant difference. Here a much more refined search is performed compared with using domain randomization. The same tendency is seen with the optimization using the ADD/I metric. 

\begin{table}[tb]
\begin{center}
    \caption{The found values from the continuous parameter search compared with the heuristic parameters.}
        \label{tab:found_con}
        
\begin{tabular}{|c|l|l|l|l|l|l|l|}
    \hline 
         Param & $vt$ & $rd$ & $id$ & $is$ & $bd$ & $ad$ & $sr$ \\ \hline
         Heur. & 0.95 & 10.0 & 2.5 & 2.0 & 10 & 5 & 72 \\ \hline
         No DR Opt. & 0.27 & 17.03 & 1.24 & 2.25 & 21 & 1 & 108 \\ \hline
         
         
         Opt. ADD & 0.275 & 12.85 & 0.77 & 3.49 & 59 & 5 & 66 \\ \hline
         Opt.     & 0.174 & 19.88 & 4.85 & 1.24 & 86 & 12 & 108 \\ \hline
    \end{tabular}
\end{center}
\vspace{-6mm}
\end{table}

\subsection{Discrete Optimization}

The discrete optimization is performed with the parameters found in the continuous optimization. A grid search is then performed using varying discrete parameters. The tested values are shown in Tab.~\ref{tab:tested_param}. As unfeasible parameter settings cannot be tested, e.g., $PE=10$ when $PC=8$, the final test iterations are 576.
As the optimization is performed with 15 objects, the run-time will not match the OCCLUSION dataset, which has only eight objects. 
To compute the expected run-time, the 576 iterations from the discrete optimization are used to solve Eq.~\ref{eqn:runtime} using least-square optimization. The resulting time contribution by increasing each discrete parameter is shown in Tab.~\ref{tab:part_time}.

\begin{table}[htb]
    \centering
    \caption{Values tested in the discrete grid search.}
    \label{tab:tested_param}
    \begin{tabular}{|l|l|l|l|l|}
        \hline 
             $PC$ & $PE$ & $RI$ & $DC$ & $II$ \\ \hline
             8,16,32 & 2,4,6,8,10 & 500,1500,2500 & 1,2,5,10 & 10,30,50 \\ \hline
        \end{tabular}
\vspace{-4mm}
\end{table}

\begin{table}[htb]
    \centering
    \caption{Time consumption of each part of the model.}
    \label{tab:part_time}
    \begin{tabular}{|c|c|c|c|c|c|}
        \hline
        Part & $t_{pre}$ & $t_{net}$ & $t_{ran}$ & $t_{depth}$ & $t_{icp}$ \\
        \hline
        Time (s) & 8.57e-01 & 7.99e-03 & 2.70e-04 & 9.12e-03 & 1.67e-04 \\ \hline
    \end{tabular}
    \vspace{-2mm}
\end{table}

\begin{figure}[htb]
    \vspace{1.5mm}
    \centering
    \includegraphics[trim=0 0 0 35, clip, width=0.95\linewidth]{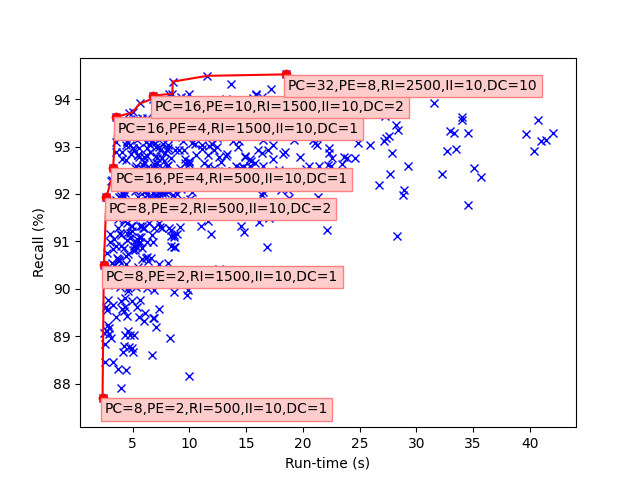}
    \caption{ The output of the discrete optimization shown as blue crosses. The red line shows the best-performing parameter sets in relation to run-time. Parameter values are visualized for some of the parameters which have the best trade-off between run-time and performance. }
    \label{fig:disc_opt_graph}
    \vspace{-1mm}
\end{figure}

The results are sorted according to run-time and performance, and the results are shown in Fig.~\ref{fig:disc_opt_graph}. Of the 576 iterations, 14 parameter sets give the best trade-off between run-time and performance. The least influential parameter is found to be the number of C2F-ICP iterations, which is always at 10.
While the depth check improves the performance, it also significantly increases the run-time. It is, therefore, only increased in value at the maximum run-time.


\begin{table}[tb]
\begin{center}
    \caption{The resulting run-time and recall for the heuristic approach and with optimized parameters, with different run-times according to Fig.~\ref{fig:disc_opt_graph}. Results are shown with the original MASK R-CNN detector and the CosyPose detector. }
    \label{tab:final_parameters}
    \small
\begin{tabular}{|l|l|l|}
    \hline
        Method & Run-time (s) & Recall ($\%$) \\ \hline
        Previous method & 3.5 & 77.2 \\ \hline
        Heuristic $<$4 sec (prev. net) & 3.4 & 77.58 \\ \hline
        Heuristic $<$4 sec & 3.4 & 77.63 \\ \hline
        
        No DR Opt. $<$4 sec & 3.7 & 77.40  \\ \hline
        ADD/I Opt. $<$4 sec & 3.6 & 77.49 \\ \hline
        
        Opt. $<$4 sec, MASK R-CNN &  3.6 & 78.91 \\ \hline
        Opt. Max, MASK R-CNN & 26.3 & 80.11 \\ \hline
        Opt. $<$4 sec, CosyPose & 3.6  & 80.45 \\ \hline
        Opt. Max, CosyPose & 26.2 & 82.03 \\ \hline
    \end{tabular}
\end{center}
\vspace{-6mm}
\end{table}

\begin{table*}[htb]
    \vspace{1.5mm}
    \centering
    \caption{ Results for the LINEMOD dataset \cite{hinterstoisser2012model} in \% recall with the ADD/I score. The competing methods are DPOD \cite{zakharov2019dpod}, SSD-6D \cite{kehl2017ssd} (obtained from \cite{wang2019densefusion}), DenseFusion \cite{wang2019densefusion}, PointVoteNet \cite{hagelskjaer2019pointvotenet}, PVN3D \cite{he2020pvn3d} and the previous method \cite{hagelskjaer2020bridging}. Rotation invariant objects are marked with an *. }
    \label{tab:linemod}

    \small
\begin{tabular}{|l|c|c|c|c||c|c|c|c|c|}
\hline
  \begin{tabular}{@{}c@{}} Training \\ Data \end{tabular} & \multicolumn{4}{c||}{\textbf{Real}} & \multicolumn{5}{c|}{\textbf{Synthetic}}       \\
  \hline
   & DenseFusion & DPOD & PointVoteNet & PVN3D & DPOD & SSD6D & Prev. & \multicolumn{2}{c|}{ParaPose} \\
   &  &  &  &  &  &  & & $<4s$ & MAX \\

        \hline
        Ape        & 92      & 87.7       & 80.7 &  97.3  & 55.2      & 65    & 97.7 & 97.9 & 97.9 \\
        Bench v.   & 93      & 98.5       & 100  &  99.7  & 72.7      & 80    & 99.8 & 100 & 100 \\
        Camera     & 94      & 96.1       & 100  &  99.6  & 34.8      & 78    & 98.3 & 97.9 & 97.9 \\
        Can        & 93      & 99.7       & 99.7 &  99.5  & 83.6      & 86    & 98.8 & 99.9 & 99.9 \\
        Cat        & 97      & 94.7       & 99.8 &  99.8  & 65.1      & 70    & 99.9 & 99.9 & 99.8 \\
        Driller    & 87      & 98.8       & 99.9 &  99.3  & 73.3      & 73    & 99.2 & 100 & 100 \\
        Duck       & 92      & 86.3       & 97.9 &  98.2  & 50.0      & 66    & 97.8 & 98.3 & 98.3 \\
        Eggbox*    & 100     & 99.9       & 99.9 &  99.8  & 89.1      & 100   & 97.7 & 98.5 & 98.5 \\
        Glue*      & 100     & 96.8       & 84.4 &  100   & 84.4      & 100   & 98.9 & 99.9 & 100 \\
        Hole p.    & 92      & 86.9       & 92.8 &  99.9  & 35.4      & 49    & 94.1 & 98.8 & 99.5 \\
        Iron       & 97      & 100        & 100  &  99.7  & 98.8      & 78    & 100  & 100 & 100  \\
        Lamp       & 95      & 96.8       & 100  &  99.8  & 74.3      & 73    & 92.8 & 100 &  99.9 \\
        Phone      & 93      & 94.7       & 96.2 &  99.5  & 47.0      & 79    & 99.1 & 99.3 & 99.4 \\
        \hline
        Average    & 94.3    & 95.15      & 96.3  & \textbf{99.4}  & 66.4      & 79    &  98.0 & 99.3 & 99.3 \\
        \hline
        \end{tabular}
        \vspace{-4mm}
\end{table*}

\subsection{Performance on real data}

Several experiments are performed to test the effectiveness of the synthetic parameter optimization method. All tests are performed on the OCCLUSION \cite{brachmann2014learning} dataset. 
The tests are performed with the parameters found in the parameter optimization. For the discrete parameters, two-parameter sets are selected, one for a run-time under four seconds and one with maximum run-time. The parameters are ($PC=16,PE=4,RI=1500,II=10,DC=1$) and ($PC=32,PE=8,RI=2500,II=10,DC=10$), respectively. 

We also test the previously trained network with our modified method, without the parameter optimization but with the same discrete parameters. Additionally, we test the new network with heuristic parameters and with parameter optimization performed without domain randomization. 
For the optimized parameters, tests are performed both with the MASK R-CNN detector and the CosyPose detector.
The resulting recall and run-times of the tests are shown in Tab.~\ref{tab:final_parameters}.
The results show that the network using optimized domain randomization has the same performance as the manually tuned model but does not significantly increase the performance. The optimization without domain randomization shows poor generalization and performs worse than heuristically tuned parameters.
However, the significance of the parameter optimization with domain randomization is shown. The same method increases from a recall of 77.63~\% to 78.91~\% by tuning the parameters. 
The optimization using the ADD/I metric shows poor results, verifying the importance of the BOP metric.
The maximum run-time parameters are also shown to increase the performance, going to a recall of 80.11~\%. Additionally, using the CosyPose detector gives superior results, increasing the recall to 82.0~\%.

\subsection{Comparison with state-of-the-art methods}

\begin{table}[tb]
    \centering
    \caption{ Results on the OCCLUSION dataset \cite{brachmann2014learning} in \% recall with the ADD/I score. The compared methods are PointVoteNet \cite{hagelskjaer2019pointvotenet}, PVN3D \cite{he2020pvn3d}, PoseCNN \cite{xiang2017posecnn}, and the previous method \cite{hagelskjaer2020bridging}. The score for \cite{he2020pvn3d} is from \cite{hesupplementary}. Rotation invariant objects are marked with an *.}
    \label{tab:occlusion}
    \small
    \begin{tabular}{|l|c|c|c||c|c|c|}
  \hline
   \begin{tabular}{@{}c@{}} Training \\ Data \end{tabular} & \multicolumn{3}{c||}{\textbf{Real}} & \multicolumn{3}{c|}{\textbf{Synthetic}}       \\  
    \hline
    
     & {\footnotesize \cite{hagelskjaer2019pointvotenet}} & {\footnotesize \cite{he2020pvn3d}} & {\footnotesize \cite{xiang2017posecnn}} &  {\footnotesize Prev.} & \multicolumn{2}{c|}{\footnotesize ParaPose}  \\
     &  &  &  & & {\footnotesize $<$ 4 s} & {\footnotesize MAX}  \\
         \hline
         Ape          & 70.0 & 33.9 & 76.2 & 66.1 & 66.6 & 68.5 \\
         Can          & 95.5 & 88.6 & 87.4 & 91.5 & 94.3 & 95.7 \\
         Cat          & 60.8 & 39.1 & 52.2 & 60.7 & 66.3 & 68.8 \\
         Driller      & 87.9 & 78.4 & 90.3 & 92.8 & 92.8 & 94.6 \\
         Duck         & 70.7 & 41.9 & 77.7 & 71.2 & 73.7 & 74.8 \\
         Eggbox*      & 58.7 & 80.9 & 72.2 & 69.7 & 71.3 & 73.4 \\
         Glue*        & 66.9 & 68.1 & 76.7 & 71.5 & 81.1 & 81.8 \\
         Hole p.      & 90.6 & 74.7 & 91.4 & 91.5 & 96.4 & 97.3 \\
         \hline
         Average     & 75.1 & 63.2 & 78.0 & 77.2 & \textbf{80.5} & \textbf{82.0} \\
         \hline
    \end{tabular}
    \vspace{-6mm}
\end{table}

The method is compared with current state-of-the-art methods to show the effectiveness of the developed parameter optimization. Only methods using RGB-D data are included as these have the highest performance. 
We showcase the results using the CosyPose \cite{labbe2020cosypose} detector with the optimized parameters for run-time under four seconds and maximum run-time. Results are shown for the LINEMOD and the OCCLUSION datasets. The performance on LINEMOD is shown in Tab.~\ref{tab:linemod}. The results show that our method outperforms all methods trained on synthetic data and most methods using real training data, while only PVN3D \cite{he2020pvn3d} obtain better results.
The results for the OCCLUSION dataset are shown in Tab.~\ref{tab:occlusion}. Apart from the previous method \cite{hagelskjaer2020bridging}, all methods have been trained on real data. Our methods outperforms all previous methods and obtains a recall of 82.0~\%. This is the new state-of-the-art result on this dataset, an improvement of four percentage points, and is obtained by using only synthetic data. Additionally, it should be noted that the method PVN3D \cite{he2020pvn3d} which obtained the highest score on the LINEMOD \cite{hinterstoisser2012model} dataset, only obtains a 63.2~\% recall on this more challenging dataset.

\section{CONCLUSION}
\label{conclusion}


This paper presents a method for the set-up of pose estimation using only synthetic data. The network is trained, and parameters are optimized using synthetic data. The method also provides the best trade-off between performance to a given run-time. While the run-time is estimated using simulation, it generally matches the real world and allows for a more straightforward set-up procedure. 

By using the optimized parameters, the pose estimation system outperforms the original method using manually tuned parameters. The developed method is compared with current state-of-the-art methods, which also use real training data. On the challenging OCCLUSION dataset, our method obtains a recall of 82.0~\%, which is the new state-of-the-art.

In future work, this method could be expanded to process entire scene point-clouds. The network can then be pre-trained on large datasets to create strong weights which increase generalizability. The network can also be trained for multiple objects, and the noise model can be trained for the entire scene.






{\small
\bibliographystyle{ieee_fullname}
\bibliography{root}
}


\end{document}